\documentclass[10pt,twocolumn,letterpaper]{article}

 \usepackage{cvpr}              % To produce the CAMERA-READY version
\definecolor{cvprblue}{rgb}{0.21,0.49,0.74}
\usepackage[pagebackref,breaklinks,colorlinks,allcolors=cvprblue]{hyperref}

\def\paperID{29} % *** Enter the Paper ID here
\def\confName{CVPR}
\def\confYear{2025}

\title{Secure Diagnostics: Adversarial Robustness Meets Clinical Interpretability}

\author{
Mohammad Hossein Najafi\\
Sharif University of Technology\\
{\tt\small mh.najafi@ee.sharif.edu}
\and
 Mohammad Morsali\\
Sharif University of Technology\\
{\tt\small mohammad.morseli@ee.sharif.edu}
\and
Mohammadreza Pashanejad\\
University of Tehran\\
{\tt\small Mehrab.Pashanejad@ut.ac.ir}
\and
Saman Soleimani Roudi\\
Sharif University of Technology\\
{\tt\small saman.soleimani@ee.sharif.edu}
\and
Mohammad Norouzi\\
Shahid Beheshti University of Medical Sciences\\
{\tt\small ar.jahangiri@mail.sbu.ac.ir}
\and
Saeed Bagheri Shouraki\\
Sharif University of Technology\\
{\tt\small bagheri-s@sharif.edu}
}

\usepackage[margin=1in]{geometry}
\usepackage{graphicx}
\usepackage{multirow} % For multirow in tables
\usepackage{booktabs} % Optional: nicer table rules

\begin{document}
\maketitle
\begin{abstract}
Deep neural networks for medical image classification often fail to generalize consistently in clinical practice due to violations of the i.i.d. assumption and opaque decision-making. This paper examines  interpretability in deep neural networks fine-tuned for fracture detection by evaluating model performance against adversarial attack and comparing interpretability methods to fracture regions annotated by an orthopedic surgeon. Our findings prove that robust models yield explanations more aligned with clinically meaningful areas, indicating that robustness encourages anatomically relevant feature prioritization. We emphasize the value of interpretability for facilitating human-AI collaboration, in which models serve as assistants under a human-in-the-loop paradigm: clinically plausible explanations foster trust, enable error correction, and discourage reliance on AI for high-stakes decisions. This paper investigates robustness and interpretability as complementary benchmarks for bridging the gap between benchmark performance and safe, actionable clinical deployment.  
\end{abstract}    
\section{Introduction}
\label{sec:intro}
Deep neural networks (DNNs) have revolutionized image recognition, achieving superhuman performance on benchmark datasets. This success has accelerated their adoption in high-stakes domains where accurate and timely predictions are critical. Among these, medical imaging stands out due to its profound impact on patient care. AI-driven diagnostic tools are now integrated into clinical workflows in radiology, pathology, and other specialties, assisting medical professionals in analyzing complex images ~\cite{wang2025deep, Rawat2023AIIH, Nia2023EvaluationOA}.  

However, the integration of AI in medical decision-making is not without challenges. A fundamental issue is the inherent complexity of medical data. Unlike standard machine learning benchmarks that assume independently and identically distributed (i.i.d.) data, medical data inherently deviate from the standard i.i.d. assumption: patient demographics, imaging protocols, and device-specific parameters vary widely across hospitals and regions, yielding a realistic but highly non-stationary learning environment~\cite{Cao2022ShallowAD}. This variability complicates model generalization, making it essential to design AI systems that remain robust across diverse clinical settings. While many medical studies focus primarily on achieving high detection accuracy and benchmarking against alternative methods ~\cite{mousa2024enhancingskincancerdiagnosis,yoooo,Ma2024,Ekman2025}, relatively few pay sufficient attention to model robustness—a factor equally vital for ensuring reliability and generalizability in clinical applications.

Beyond the non-i.i.d. challenge, another critical issue is the interpretability of AI models. In medical imaging, clinicians rely on AI not only for accurate predictions but also for transparent reasoning that aligns with their expertise. This highlights the growing need for explainable AI (XAI) approaches that offer interpretable visualizations, ensuring that AI-generated insights are both accurate and meaningful for clinical decision-making~\cite{Raif2023ExplainableA, Mariappan2024ExtensiveRO}. XAI techniques are gaining importance in creating trustworthy AI solutions within healthcare~\cite{Raif2023ExplainableA}. By offering transparency, accountability, and traceability, XAI helps interpret predictions or decisions made by AI-based systems in healthcare applications, including medical diagnosis and decision support systems~\cite{Mariappan2024ExtensiveRO}.

Interpretability methods can be categorized along three key dimensions~\cite{adadi2018peeking}:

\begin{itemize} \item \textbf{Interpretability Complexity}: Distinguishes between intrinsic models, which inherently offer transparency, and post-hoc methods that apply explanation techniques after training. \item \textbf{Explanation Scope}: Defines the scope of explanation as either global (analyzing the behavior of the model as a whole) or local (explaining individual predictions). \item \textbf{Model Dependence}: Addresses whether techniques are specific to a particular model architecture (dependent) or applicable across different models (agnostic). \end{itemize}

In the medical domain, post-hoc explanation techniques offer significant advantages by enhancing transparency. They allow complex models to be interpreted after training, without compromising performance~\cite{jin2023generating,retzlaff2024post}. This flexibility enables the application of sophisticated algorithms while still providing clear insights into their decision-making processes, which is crucial for clinician trust and regulatory compliance. Furthermore, local explanation techniques offer patient-specific insights, focusing on individual predictions. By elucidating the factors influencing specific diagnoses or treatment recommendations, these methods foster improved communication between healthcare providers and patients, aiding in shared decision-making and enhancing patient understanding. The integration of both post-hoc and local explanation techniques into medical practice ensures that AI models are not only powerful but also interpretable, ultimately contributing to more effective and patient-centered healthcare.
For scenarios where local, post-hoc explanations are required, interpretability maps serve as an invaluable tool. An interpretability map is a visual or structured representation that identifies and highlights the specific portions of the input data that have the greatest influence on a model's prediction. By directly linking input features to decision outcomes, these maps demystify the internal workings of complex models, thereby enhancing transparency and supporting more informed clinical decisions.
Although interpretability methods can help decision-making, measuring them remains a challenge due to their inherently subjective nature~\cite{doshivelez2017rigorousscienceinterpretablemachine}. Current evaluation approaches often require subjective input from humans or incur high computational costs with automated evaluation~\cite{Lin2020WhatDY}. Interpretability evaluations can be categorized into three main types~\cite{doshivelez2017rigorousscienceinterpretablemachine}:

\begin{itemize}
\item \textbf{Application-grounded}: Evaluations where domain experts assess whether AI-generated explanations support real-world decision-making. \item \textbf{Human-grounded}: Simplified tests where laypersons evaluate AI explanations based on comprehensibility. \item \textbf{Functionally-grounded}: Automated assessments based on mathematical or computational metrics. \end{itemize}

In clinical settings, application-grounded evaluation is often preferred. This approach involves domain experts assessing whether explanations meaningfully contribute to decision-making~\cite{gambetti2025surveyhumancenteredevaluationexplainable, amarasinghe2023importanceapplicationgroundedexperimentaldesign}. It aligns with human-computer interaction principles, where explanations are evaluated based on their impact on clinical tasks such as error identification, insight discovery, or bias reduction~\cite{doshivelez2017rigorousscienceinterpretablemachine}.

In addition to interpretability, adversarial robustness is essential for AI systems in medical imaging. Even slight adversarial perturbations can trigger misdiagnoses or delays in detecting life-threatening conditions, significantly compromising patient safety. Despite strides in predictive accuracy, many models remain vulnerable to these subtle attacks ~\cite{bortsova2021adversarial}. Moreover, while research often prioritizes accuracy, fewer studies address robustness . This concern is heightened with the rise of domain-specific foundation models in medical imaging (XFMs), which, although enhancing diagnostics, are also prone to adversarial attacks that can lead to critical misclassifications ~\cite{raj2025examiningthreatlandscapefoundation, G.Wen_BrendaRodriguez-Niño_FurkanY.Pecen_D.Vining_N.Garg_M.Markey_2017}.

 The imperative for AI in medical imaging to prioritize interpretability, robustness and trustworthiness, rather than raw accuracy alone, naturally aligns with human-in-the-loop (HITL) machine learning \cite{holzinger2016interactive}, which integrates clinicians as essential collaborators to validate AI outputs, provide contextual expertise, and mitigate risks like bias or over-reliance on automation. By framing AI as an assistant rather than an autonomous decision-maker, HITL ensures human oversight, enhancing diagnostic reliability and safety while addressing ethical and practical challenges.

Motivated by the challenges discussed above, we present a systematic study to examine how interpretability and robustness interact. Specifically, we:
\begin{enumerate}
    \item \textbf{Fine-tune Robust DNNs:} Fine-tune multiple robust DNN architectures (pretrained on ImageNet) on the Bone Fracture X-ray dataset~\cite{Rodrigo2021}, incorporating images from multiple clinical sources.
    
    \item \textbf{Rank Model Robustness:} Sort the models fine-tuned on the bone fracture dataset by evaluating their adversarial resilience under $\ell_\infty$-bounded perturbations.
    
    \item \textbf{Evaluate Interpretability:} Assess model interpretability via multiple interpretability maps using an application-grounded approach. In this approach, a practicing radiologist first provided qualitative comparisons between the interpretability maps, assessing which visualizations best align with clinical reasoning patterns. Subsequently, the radiologist independently annotated the clinically relevant regions for fracture diagnosis in the medical images, and these annotations were used for a quantitative comparison between the maps.
\end{enumerate}

\section{Related work}
\label{sec:related_work}
In this work, we utilize interpretability maps as post-hoc, local interpretability methods to elucidate the decision-making processes of DNNs. Accordingly, our review is structured as follows: the first subsection delves into various interpretability maps; the second subsection explores their historical application in medical imaging; and the final subsection examines the relationship between interpretability and robustness.

\subsection{Interpretability Maps as an Explainability Technique for DNNs}

DNNs are often criticized for their black-box nature, which obscures the rationale behind their predictions. To address this, interpretability maps have been developed as pivotal tools to visualize and understand the features influencing model decisions.
One of the earliest methods, saliency maps~\cite{simonyan2013deep}, computes the gradient of the output class score with respect to the input features, highlighting regions that most significantly affect the model's prediction. This approach is computationally efficient, requiring only a single backward pass, and provides a straightforward visualization of feature importance. However, saliency maps can produce noisy and less interpretable visualizations and are sensitive to minor input perturbations, which may affect their reliability.
Building upon this, Occlusion method systematically masks portions of the input data to observe changes in the model's output, thereby identifying regions critical ~\cite{zeiler2014visualizing}. These techniques are intuitive and directly assess the impact of specific input regions on the model's decision. Yet, they can be computationally intensive due to the need for multiple forward passes, and the choice of occlusion pattern (\eg, shape and size of the masked region) can influence results.
To enhance interpretability, Gradient-weighted Class Activation Mapping (Grad-CAM) combines gradient information with feature maps from the last convolutional layer to produce class-discriminative interpretability maps~\cite{selvaraju2017grad}. This method effectively identifies class-specific regions in input images and is applicable to a wide range of convolutional neural network (CNN) architectures. Nonetheless, Grad-CAM generates coarse maps due to the low resolution of deeper convolutional layers and is less effective for models lacking convolutional structures.
Another approach, Integrated Gradients~\cite{sundararajan2017axiomatic}, attributes the prediction by integrating gradients along a path from a baseline input to the actual input. This method provides more accurate attribution by considering the entire path from baseline to input and is theoretically grounded with desirable properties for attribution methods. However, the choice of baseline can significantly influence results and may not be straightforward, and the method is computationally demanding due to the need for multiple gradient computations along the path.
Deep Learning Important FeaTures (DeepLIFT)~\cite{shrikumar2017learning} offers an alternative by comparing the activation of each neuron to a reference activation and assigning contribution scores based on the differences. This approach offers clear explanations by comparing activations to a reference state and is efficient, requiring only a single backward pass for computation. Yet, selecting an appropriate reference state can be challenging and may affect interpretability, and implementation can be complex due to the need for customized backpropagation rules\cite{hosseini2025ultraunveilinglatenttoken}.
These interpretability methods serve as essential tools for decoding the complex decision-making processes of DNNs, each with its unique strengths and limitations. Understanding these techniques enables practitioners to select the most suitable approach based on their specific application needs and computational constraints.

\subsection{Interpretability Maps in Medical Imaging}
Interpretability maps have been applied in medical imaging in recent years to address the critical need for transparency in clinical decision-making. Early work by Sayres \etal~\cite{sayres2019using} demonstrated the potential of these methods by employing Integrated Gradients to highlight influential regions in retinal images, thereby assisting clinicians in understanding deep neural network predictions for diabetic retinopathy grading and effectively bridging AI outputs with clinical insights. Building on this foundation, subsequent research expanded these techniques to urgent diagnostic challenges; for example, during the COVID-19 pandemic, Panwar \etal~\cite{panwar2020deep} applied Grad-CAM to visualize critical regions in chest X-rays and CT scans, which enhanced trust in rapid diagnostic systems. Simultaneously, Aggarwal \etal~\cite{aggarwal2020towards} introduced trainable saliency maps aimed at improving reliability and transparency, while Monroe \etal~\cite{monroe2021hiho} proposed a hierarchical occlusion approach designed for real-time interpretability in IoT-based medical systems.
% As the field progressed, studies began scrutinizing the reliability of these techniques. Research by Saporta \etal~\cite{Saporta2021.02.28.21252634} raised concerns that saliency maps do not always pinpoint diagnostically relevant areas, a limitation further examined by Arun \etal~\cite{arun2020assessing} through evaluations against localization benchmarks. Complementing these findings, Xia \etal~\cite{xiao2021visualization} extended the application of Grad-CAM to segmentation tasks, demonstrating its versatility in confirming that deep learning models focus on clinically pertinent regions. Further efforts shifted toward clinical validation and the integration of interpretability into model training; for instance, Ayhan \etal~\cite{ayhan2022clinical} conducted clinical assessments in ophthalmology to confirm the utility of saliency maps, while Mahapatra \etal~\cite{mahapatra2022interpretability} incorporated interpretability into training processes through inductive bias to develop inherently transparent models. In a similar vein, Kawai \etal~\cite{kawai2022compensated} refined Integrated Gradients for EEG analysis, thereby broadening the applicability of these methods across various medical modalities.
More recent evaluations have delved deeper into performance and trustworthiness, with Zhang \etal~\cite{zhang6revisiting} revisiting saliency approaches in radiology by proposing new metrics to enhance clinical adoption and Suara \etal~\cite{suara2023grad} exploring the explanatory power of Grad-CAM by weighing its benefits and limitations. Additionally, work by Jin \etal~\cite{jin2023generating} leveraged DeepLIFT to generate post-hoc explanations for multi-modal image analysis, while a comprehensive survey by Patr{\'\i}cio \etal~\cite{patricio2023explainable} synthesized diverse techniques to highlight their collective role in enhancing AI transparency.
While these studies advanced the use of interpretability maps in medical imaging, they primarily employed standard (non-robust) deep learning models. None of these works explicitly investigated the impact of adversarial robustness on the reliability of interpretability maps. For instance, they did not explore whether robust models produce more clinically plausible explanations compared to standard models. This gap highlights the need to study how robustness influences interpretability in safety-critical medical applications.
\subsection{Interpretability and Adversarial Robustness}
Adversarially robust models, designed to resist intentional perturbations, have become more interpretable by focusing on semantically meaningful features. Ross and Doshi-Velez ~\cite{ross2018improving} introduced a regularization method that aligns a model's input gradients with features that humans find significant, thereby clarifying its decision process. Similarly, Etmann \etal~\cite{etmann2019connection} showed that robust models generate saliency maps closely aligned with human intuition, and Zhang \etal~\cite{zhang2019theoretically} demonstrated that adversarial training leads to feature representations that are easier for humans to understand. Therefore, enhancing robustness not only improves defense against attacks but also makes the model's internal logic more transparent, which is critical in applications such as medical imaging.

\section{Methodology}
\label{sec:methodology}

Robust models truly learn human-relevant features, and their maps on X-rays focus on key anatomical structures (the bones and fracture lines). We leverage this insight to choose the models based on their robustness. By comparing maps across different models, we can assess how robustness affects the model's focus and whether it maintains human-like attention. We expect a well-behaved fracture robust detector to highlight the bone regions, mainly the fracture site, rather than arbitrary background pixels. So, we use state-of-the-art adversarially robust models from the RobustBench, which provides pre-trained ImageNet classifiers known for high robustness. These models originally output 1000 classes (ImageNet labels); we adapt them to our binary fracture detection task by replacing the final linear layer with a new one with two outputs (fractured vs. healthy). 

To exploit the robustness of the pre-trained models without distorting their learned representations, we employ transfer learning with frozen feature layers. In practice, we freeze all convolutional and intermediate layers and only train the newly added final layer (the fully connected classifier) on the fracture dataset, a common approach in medical imaging tasks where pre-trained ImageNet models are fine-tuned on limited data. We preserve the model’s adversarially robust features by not updating the backbone weights while only the last layer’s
weights are initialized for our two-class problem.
 This approach mitigates overfitting and maintains the robustness prior to the original training.

We evaluate robustness using a standard attack. First, we implement the Project gradient descent(PGD) \cite{madry2019deeplearningmodelsresistant} as our adversarial attack. We employ this single adversarial attack to obtain the robustness order of our six robust models from the Robustbench rank, which was fine-tuned with our dataset. After performing the attack, we sorted the robust models by comparing the model's performance under a PGD attack with $\epsilon = 4/255$ (threshold)  shown in \cref{tab:combined}. The order of model performance in our dataset was approximately adapted to the ranking of robustbench models. 

We compare each model's accuracy on unperturbed test images (clean accuracy) and adversarially attacked images (adversarial accuracy). A small drop in accuracy under attack indicates that the model successfully resisted the perturbations, whereas a large drop means that the adversarial noise significantly fooled the model. We quantify performance degradation using the metric \(\Delta \text{Acc} = \text{Acc}_{clean} - \text{Acc}_{adv}\). This accuracy drop percentage captures the loss of performance due to the attack.

\begin{table*}[!ht]
\centering
\caption{Mapping between RobustBench Rank, Performance Rank, and comparison of model model performance under a PGD attack with $\epsilon = 4/255$.}
\label{tab:combined}
\resizebox{\textwidth}{!}{%
\begin{tabular}{lccccc}
\toprule
\textbf{Model} & \textbf{Performance Rank} & \textbf{RobustBench Rank} & \textbf{Test Accuracy (\%)} & \textbf{PGD (4/255) (\%)} & \textbf{Delta Acc (\%)} \\
\midrule
MeanSparse \cite{amini2024meansparseposttrainingrobustnessenhancement}     & 1 & 6  & 99.21 & 86.96 & 12.25 \\
Swin \cite{liu2023comprehensivestudyrobustnessimage}                      & 2 & 2  & 97.63 & 82.21 & 15.42 \\
NIG \cite{rodríguezmuñoz2024characterizingmodelrobustnessnatural}         & 3 & 15 & 97.63 & 79.45 & 18.18 \\
Revisiting \cite{singh2023revisitingadversarialtrainingimagenet}           & 4 & 12 & 99.01 & 78.85 & 20.16 \\
Light \cite{debenedetti2023lightrecipetrainrobust}                       & 5 & 20 & 98.62 & 69.37 & 29.25 \\
Standard \cite{he2016deep}                                              & 6 & 30 & 93.48 & 0     & 93.48 \\
\bottomrule
\end{tabular}%
}
\end{table*}

After evaluating the models, we generate Saliency, Occlusion, and DeepLIFT maps for each model's predictions to understand where the models focus their attention.

For saliency map we use gradient-based saliency techniques, we calculate the gradient of the predicted class score for each input pixel. These gradients highlight image regions most strongly influencing the model's prediction (higher gradient magnitude = more significant influence).

The saliency map $\mathbf{S}(\mathbf{x})$ for an input image $\mathbf{x}$ is computed as:
\begin{equation}
    \mathbf{S}(\mathbf{x}) = \left| \frac{\partial f_c(\mathbf{x})}{\partial \mathbf{x}} \right|
    \label{eq:sensitivity}
\end{equation}

where $f_c(\mathbf{x})$ is the model's output logit for class $c$.

For RGB images, we take the maximum across channels:
\begin{equation}
    \mathbf{S}_{\text{final}}(\mathbf{x}) = \max_{k \in \{R,G,B\}} \left| \frac{\partial f_c(\mathbf{x})}{\partial \mathbf{x}_k} \right|
    \label{eq:final-sensitivity}
\end{equation}

Occlusion sensitivity mapping is an interpretability technique. The idea is to systematically “mask out” (occlude) different parts of an input image and observe how the model’s output (\eg, classification confidence or accuracy) changes. In practice, a small patch (\eg, a square window) is moved across the image; at each position, the patch region is replaced with a baseline value (such as a neutral gray or zero), and the model’s prediction is recorded. The areas causing the highest accuracy loss (largest $\Delta Acc$) are considered the most critical to decision-making. 

We define the occlusion-based sensitivity of a model prediction as follows. For an input image \(x\) and target class \(c\), let \(x^{(i,j)}\) denote the image with a patch at position \((i,j)\) occluded. Then the sensitivity at \((i,j)\) is
\begin{equation}
    S(i,j) = f(x)_c - f(x^{(i,j)})_c
    \label{eq:saliency-diff}
\end{equation}

where \(f(x)_c\) is the model output for class \(c\). This score measures how much the occlusion at \((i,j)\) affects the prediction.
To recap the sensitivity across all \(C\) channels (\eg, RGB), we define
\begin{equation}
    S_{\text{total}}(i,j) = \sum_{k=1}^{C} S_k(i,j)
    \label{eq:total-saliency}
\end{equation}

Sensitivity can be evaluated at intervals with a stride $(s_h, s_w)$ to reduce computation.
\begin{equation}
    S_{\text{stride}}(i,j) = S(i \cdot s_h, j \cdot s_w)
    \label{eq:stride-saliency}
\end{equation}

For visualization purposes, sensitivity values are normalized to the range \([0,1]\) by
\begin{equation}
    S_{\text{norm}}(i,j) = \frac{S(i,j) - \min(S)}{\max(S) - \min(S)}
    \label{eq:normalized-saliency}
\end{equation}

Lastly, occlusion sensitivity is closely related to gradient-based techniques. Indeed, it is possible to approximate it.
\begin{equation}
    S(i,j) \approx -\nabla f(x)_c \cdot \Delta x_{i,j}
    \label{eq:gradient-approx}
\end{equation}

which shows that the sensitivity is proportional to the negative inner product of the gradient of \(f(x)_c\) with respect to \(x\) and the change in the input at \((i,j)\). 
 
 DeepLIFT (Deep Learning Important Features) is a gradient-based attribution method that explains by assigning an importance score to each input feature (pixel) based on the network's output. In essence, DeepLIFT backpropagates contribution scores through the network: each neuron's contribution is distributed to its inputs, and the score for every input pixel reflects how changing that pixel from a reference value leads to a change in the output. By choosing a valid reference (often a blurred or empty image), DeepLIFT bypasses zero-gradient saturation issues and can capture non-linear effects missed by simple gradients.

 Consider an input image \( x \in \mathbb{R}^{H \times W \times C} \), where \( H \) is the height, \( W \) is the width, and \( C \) is the number of channels. The baseline (or reference input) is defined as the zero tensor ($x_{\text{ref}} = \mathbf{0}$). We want to detect the fracture class, denoted by the target class \( t = 0 \). The model's output logit for class \( t \) is given by \( f_t(x) \).

For each pixel \((i,j,c)\) (with \( c \) indicating the channel), we compute the contribution of the change in the input to the change in the model output as follows:
\begin{equation}
    C_{\Delta x_{i,j,c} \Delta f_t} = \frac{\Delta f_t}{\Delta x_{i,j,c}} \cdot \bigl( x_{i,j,c} - x_{\text{ref},i,j,c} \bigr)
    \label{eq:correction-term}
\end{equation}

where
\begin{equation}
    \Delta f_t = f_t(x) - f_t(x_{\text{ref}}) = f_t(x) 
    \label{eq:function-difference}
\end{equation}

( \eqref{eq:function-difference} since \( f_t(0) = 0 \)) and
\begin{equation}
    \Delta x_{i,j,c} = x_{i,j,c} - 0 = x_{i,j,c}
    \label{eq:delta-x}
\end{equation}

The per-pixel saliency is then obtained by taking the maximum absolute contribution over the RGB channels:
\begin{equation}
    D(i,j) = \max_{c \in \{R,G,B\}} \left| C_{\Delta x_{i,j,c} \Delta f_t} \right|
    \label{eq:max-correction-term}
\end{equation}

Integrated Gradients (IG) attributes model predictions to input features by computing the path integral of gradients between a baseline (\eg, neutral reference input) and the actual input. This technique satisfies two fundamental axiomatic properties: \textit{Sensitivity} (features with zero contribution receive no attribution) and \textit{Completeness} (attributions sum to the difference between model outputs at the input and baseline). The attribution for the \(i\)-th feature is formally defined as:
\begin{equation}
    \text{IG}_i(x) = (x_i - x'_i) \times \int_{0}^{1} \frac{\partial F(x' + \alpha(x - x'))}{\partial x_i} \, d\alpha
    \label{eq:integrated-gradients}
\end{equation}

where \(x\) is the input, \(x'\) the baseline, and \(F\) the model function. Our implementation approximates this integral through a 20-step Riemann sum (\(n\_steps\)), with either zero or mean-shifted baselines.

 \begin{figure*}[ht]
    \centering
    \includegraphics[width=0.8\textwidth]{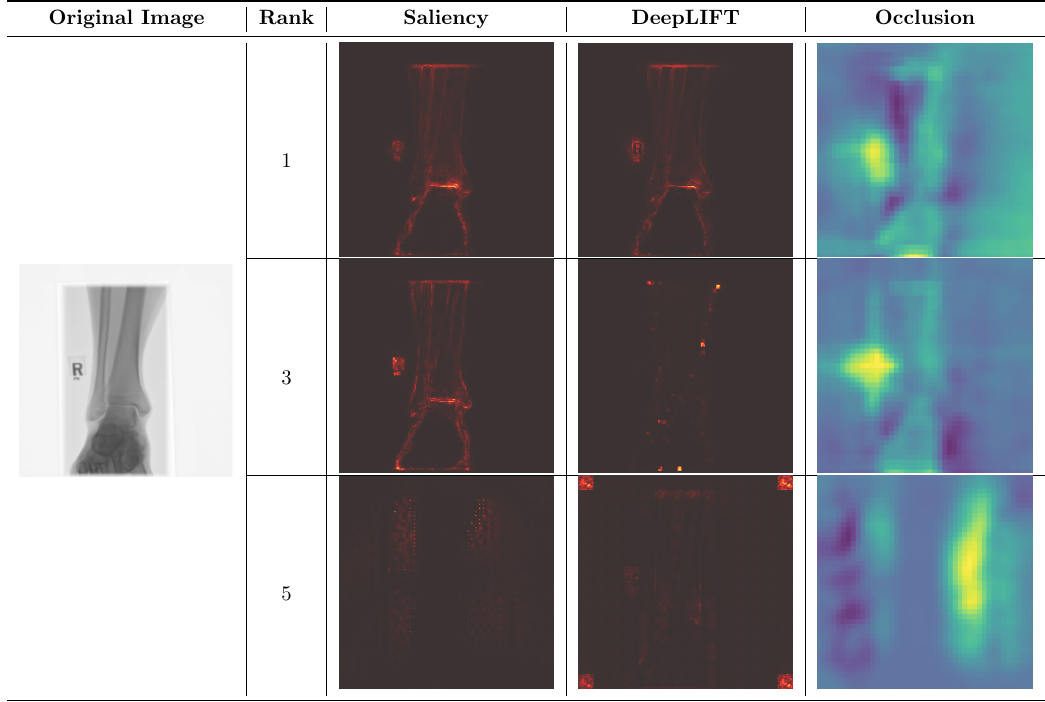}
    \caption{Comparison of interpretability maps (Saliency, DeepLIFT, Occlusion) 
    for three robust models (ranks 1, 3, 5 in our comparison of model performance in \cref{tab:combined}) alongside the original image. The higher-ranked robust model concentrates more on the fracture site, 
    while the lower-ranked model’s focus is broader and less clinically aligned.}
    \label{fig:comparison_maps}
\end{figure*}

\section{Evaluation}
  \label{sec:experiments}

We conducted our experiments on the Bone Fracture Multi-Region X-ray dataset from Kaggle. This dataset contains a total of 10,580 radiographic images labeled as either fractured or non-fractured, covering various body regions (\eg, upper limbs, lower limbs, hips, knees, spine). We used the provided training set of 9,246 images for model fine-tuning, a validation set of 828 images for hyperparameter tuning, and a test set of 506 images for final evaluation (fracture vs. healthy cases are approximately balanced in each split). All images were preprocessed (resized and normalized) before being fed to the models. Because only the final linear layer’s weights were updated (with all other layers frozen as described earlier), training was relatively fast and not prone to overfitting.

All robust model backbones were loaded from RobustBench’s library of pre-trained models. Each model (\eg, an adversarially trained ResNet) comes from prior work and has documented robustness on ImageNet. We transfer their adversarially learned features to our task using these as initialization. We ensured a fair comparison across all models by using consistent attack parameters, specifically a PGD attack with 10 steps, a step size of 1/255, and $\epsilon$ of 4/255.

The introduced models from the robust bench were fine-tuned over 30 epochs using the Kaggle P100 GPU. We utilized the Adam optimizer with a learning rate of 0.001, and the loss function employed was cross-entropy. The batch size for this training process was set to 32.

Next, we evaluate the maps of the models in the X-ray images to interpret their decision-making. We inspect whether the highlighted regions coincide with the actual fracture or relevant bone structures for correctly classified fracture cases. We observed that models with adversarial robustness tend to have maps concentrated on the fracture site or the affected bone. In contrast, less robust sometimes highlights irrelevant regions. This qualitative assessment is vital in a medical context: a model that focuses on the correct area (\eg, the crack in the bone) is more trustworthy than one that makes a decision based on it. Interestingly, even when adversarial noise is added, just models often retain focus on important anatomical regions, while the attention of a non-robust model shifts or becomes erratic. These observations align with the literature that robust models have more meaningful and human-aligned explanations.

 The following parts detail the results of this evaluation, with comparisons of accuracy and visual explanations backed by references to established findings in robustness research and medical AI.
All maps can be used to evaluate how well the model’s attention aligns with the actual fracture site marked by radiologists or orthopedists (the ground truth).

One can verify localization using occlusion maps by observing the impact of masking the fracture area. If the model relies on area to detect the fracture, occluding it will cause a notable drop in its confidence or accuracy. This would be reflected as a bright spot on the occlusion map at the fracture location. 
In contrast, occluding other regions (like a region of healthy bone or soft tissue) might have minimal effect on the output, showing that those regions were not pivotal for the prediction.DeepLIFT maps (or other saliency methods) provide a complementary view by directly highlighting pixels important to the model. On a correctly detected fracture X-ray, a DeepLIFT map would ideally show a concentrated highlight on the fracture line – essentially tracing the model’s gaze to the break in the bone. This indicates "where" the model is looking. For instance, pixels along the crack might have high positive scores contributing to the “fracture” class.

\cref{fig:comparison_maps} illustrates a direct comparison of Saliency, DeepLIFT, and Occlusion maps for three models with different robustness ranks (1, 3, and 5). Each row shows the original X-ray image on the left, followed by the respective interpretability maps. As seen in the figure, the highest-ranked robust model in our study (Rank 1) concentrates more clearly on the actual fracture site, highlighting the bone discontinuity. In contrast, the lower-ranked model (Rank 5) exhibits a less focused attention region. This visual evidence supports our findings that stronger adversarial robustness often correlates with more meaningful interpretability.

We conducted a targeted study to investigate whether our robust classification models inadvertently learn to focus on diagnostic features like expert orthopedic specialists. Although our primary task was classification, we hypothesized that robust models might develop an internal representation that emulates the perceptual patterns of human experts, focusing on key fracture details.

Two experienced orthopedic surgeons were involved in a detailed review of the fracture images to test this hypothesis. Initially, they were provided with 69 unique fractured X-ray images (after removing rotated and duplicated images from the test dataset) and asked to annotate specific regions they considered critical for diagnosing the fracture. These annotations were then used to develop a custom metric for evaluating the precision of the model’s focus on fracture details—essentially measuring how well the saliency, occlusion, and DeepLIFT maps overlapped with the expert-marked regions.
\begin{figure}[ht]
    \centering
    \includegraphics[width=\columnwidth]{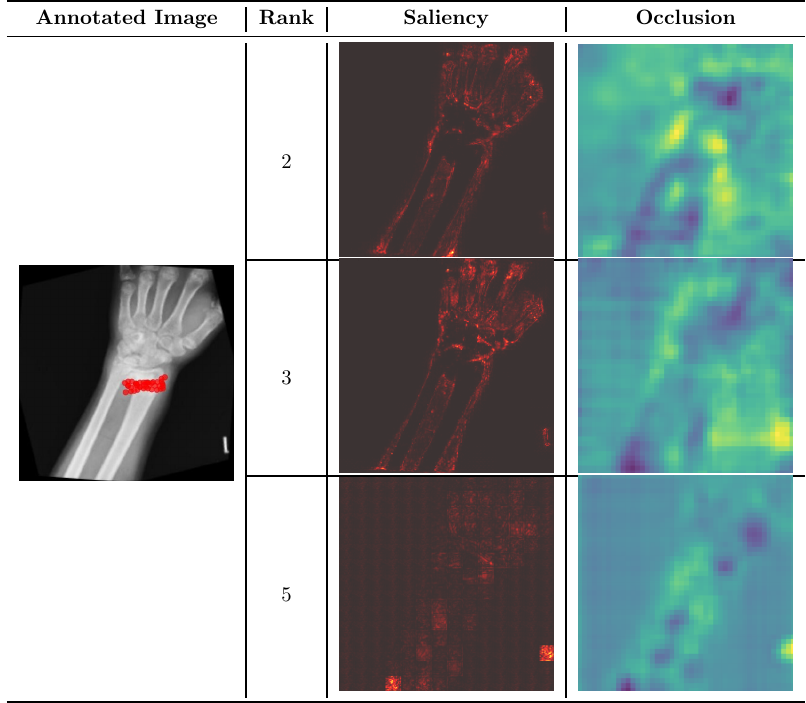}
    \caption{Comparison of specialist-annotated image with Saliency and Occlusion maps 
    for three robust models (ranks 2, 3, 5 in our comparison of model performance in Table \cref{tab:combined}) alongside the original image.}
    \label{fig:comparison_maps2}
\end{figure}

Subsequently, we presented three types of interpretability maps on a subset of ten representative images. The surgeons confirmed that the more robust models focused more on the relevant fracture features. In particular, gradient-based maps frequently highlighted the connections between bone contours and fracture lines; these maps were sensitive to structural discontinuities similar to those identified by specialists. Furthermore, the occlusion maps were observed to outperform the other methods by identifying the fracture regions, capturing the depth and extent of the break, and providing a more comprehensive visual explanation.

\cref{fig:comparison_maps2} shows an annotated X-ray image in the left column, marked by the expert as the fracture region, alongside two interpretability maps (Saliency and Occlusion) for each of the three robust models (ranks 2, 3, and 6 in Table\cref{tab:combined}). Notably, robust models accurately capture critical bone structures emphasized by the expert. Moreover, the occlusion maps offer a broader understanding of fracture extent, complementing the finer, gradient-based details visible in the saliency maps.  
\begin{table*}[ht]
\centering
\caption{Coverage overlap (\%) for three interpretability methods at different percentiles.}
\label{tab:combined_methods}
\resizebox{\textwidth}{!}{%
\begin{tabular}{c | cccc | cccc | cccc}
\toprule
\multirow{2}{*}{\textbf{Model}} & \multicolumn{4}{c|}{\textbf{Integrated Gradients}} & \multicolumn{4}{c|}{\textbf{DeepLIFT}} & \multicolumn{4}{c}{\textbf{Saliency Maps}} \\
\cmidrule(lr){2-5} \cmidrule(lr){6-9} \cmidrule(lr){10-13}
 & \textbf{95\%} & \textbf{85\%} & \textbf{75\%} & \textbf{15\%} & \textbf{95\%} & \textbf{85\%} & \textbf{75\%} & \textbf{15\%} & \textbf{95\%} & \textbf{85\%} & \textbf{75\%} & \textbf{15\%} \\
\midrule
MeanSparse \cite{amini2024meansparseposttrainingrobustnessenhancement}   & 29.11 & 57.36 & 69.64 & 96.94 & 31.59 & 60.12 & 71.12 & 97.59 & 31.65 & 62.72 & 75.96 & 98.87 \\
Swin \cite{liu2023comprehensivestudyrobustnessimage}   & 23.15 & 53.34 & 67.24 & 94.00 & 12.24 & 35.10 & 50.07 & 88.18 & 26.02 & 59.08 & 72.92 & 98.23 \\
NIG \cite{rodríguezmuñoz2024characterizingmodelrobustnessnatural}  & 20.57 & 45.74 & 57.31 & 91.36 & 18.01 & 47.69 & 62.99 & 93.95 & 20.85 & 47.99 & 64.08 & 97.53 \\
Rivisitng \cite{singh2023revisitingadversarialtrainingimagenet}  & 26.30 & 52.36 & 65.87 & 94.79 & 31.61 & 58.53 & 68.02 & 94.14 & 33.28 & 60.88 & 75.18 & 97.94 \\

Light \cite{debenedetti2023lightrecipetrainrobust} &  9.80 & 24.78 & 39.82 & 89.69 &  8.87 & 28.31 & 42.09 & 91.17 & 17.47 & 40.47 & 54.82 & 92.86 \\
Standard \cite{he2016deep}  &  3.64 & 10.59 & 20.43 & 84.38 &   N/A &   N/A &   N/A &   N/A & 10.39 & 25.84 & 37.93 & 92.76 \\
\bottomrule
\end{tabular}%
}
\end{table*}
During the evaluation, several additional insights were obtained from discussions with the experts. They highlight a potential avenue for future research, where integrating perceptual modeling could further bridge the gap between human and machine interpretations. Additionally, the specialists remarked on the improved quality of modern radiological images compared to the somewhat dated images in our dataset. They emphasized that incorporating metadata—such as imaging protocols, the presence of casts, patient age(the difference between bone of men and women or kids), or even temporal information regarding the stage of fracture healing—could significantly enhance diagnostic accuracy by providing essential contextual clues. They collected this data to detect better by observing the images.  

To make this analysis more concrete, we define a map coverage metric: the proportion of the saliency of the model (\eg, the sum of saliency values or the area of the highlighted region) that falls within the fracture region of the truth of the ground defined by experts. This metric measures how well the model’s attention aligns with the fracture. A higher saliency coverage indicates that the model focuses on the right area. This alignment between the model explanation and orthopedic is desirable for robustness and interpretability. We use this metric to compare models’ explainability quantitatively. Robust models that maintain high saliency coverage on fractures are deemed resilient and clinically interpretable – they look at the fracture when making decisions, which is precisely what a human expert would do.

 This map ensures robustness, as different attribution techniques capture varying aspects of feature 
importance. We compute pixel-level attributions for each method using the Captum frameworks\cite{kokhlikyan2020captum}, then normalize them per image and define "salient 
regions" using an adaptive percentile threshold (\eg, top 15\% of attribution values). This thresholding accounts for variability in attribution magnitude across images, thereby avoiding fixed-value biases. Coverage is calculated as the proportion of fracture points within the thresholded salient regions for a given image and its annotated fracture coordinates. 

The binary mask $\mathbf{M}$ thresholds $\mathbf{S}$ at the $\nu$-th percentile:
\begin{equation}
\mathbf{M}(x,y) = 
\begin{cases} 
1 & \text{if } \mathbf{S}(x,y) \geq \text{percentile}(\mathbf{S}, \nu), \\
0 & \text{otherwise}.
\end{cases}
\label{eq:mask-definition}
\end{equation}

Coverage metric:
\begin{equation}
\text{Point Coverage Ratio} = \frac{\left| \{(x_i, y_i) \in \mathcal{P} \mid \mathbf{M}(x_i, y_i) = 1\} \right|}{|\mathcal{P}|}
\label{eq:point-coverage-ratio}
\end{equation}

As shown in \cref{tab:combined_methods}, we report the coverage overlap (\%) at different thresholds for three interpretability methods—Integrated Gradients, DeepLIFT, and Saliency Maps—across our used models. Each threshold (5\%, 15\%, 25\%, 85\%) implies a different cutoff for highlighting significant areas in the interpretability maps. Notably, the MeanSparse\cite{amini2024meansparseposttrainingrobustnessenhancement} and Revisiting\cite{singh2023revisitingadversarialtrainingimagenet} models have more extensive coverage overlaps in all methods and imply better correspondence with the fracture areas annotated by experts. Conversely, the Standard\cite{debenedetti2023lightrecipetrainrobust} model has considerably lower coverage, especially at the more constrained thresholds (5\%  and 15\%). Also, Saliency Maps consistently have the highest coverage among the three methods, which implies they capture relevant fracture features more consistently.

This expert evaluation supports our findings that increased model robustness correlates with more human-aligned interpretability. The ability of robust models to mimic the target patterns of orthopedic experts not only builds confidence in AI-based diagnoses but also suggests the importance of employing clinical metadata.

\section{Discussion}

 Robust training constrains the model’s gradients and feature usage, often yielding more human-aligned explanations. Tsipras \etal\cite{tsipras2019robustnessoddsaccuracy} described this as an “unexpected benefit” of robustness. In contrast, standard networks often have noisy or hard-to-decipher saliency maps; robust networks’ saliency maps tend to be far more interpretable in that structures in the input image also emerge in the corresponding saliency map. Empirically, a non-robust model might technically use imperceptible pixel patterns to make a decision (hence, a raw gradient map looks like scattered noise), but a robust model is forced to rely on features stable under perturbations – typically, these are the same features a human would rely on (edges, shapes). As a result, robust model gradients and attribution maps align with salient image structures. This concept carries to fracture detection: a robust fracture detector is expected to highlight the fracture line or break more cleanly, without extraneous scatter. A slight adversarial noise is unlikely to distract the robust model’s attention to a different region; the model will continue to lock onto the actual fracture features.

 In addition, recent work has focused on improving and scaling interpretability methods to include explainability in real-time clinical decision-making. A good example is the Fast Multi-Resolution Occlusion (FMO) method, which is a perturbation-based method that significantly shortens the runtime of occlusion tests. FMO achieves an average 2.3× speedup over LIME and over 10× speedup vs\cite{Behzadi-Khormouji2021}. standard occlusion or RISE methods without sacrificing explanation quality. This efficiency (stemming from multi-scale occluding patches) makes occlusion feasible even on high-resolution images and large datasets. Another study introduced the PLI model for better explanations and with computational efficiency in mind – reporting an average inference time of 0.75 s per image vs. 1.45 s for Grad-CAM (nearly 2× faster) under the same conditions\cite{make7010012}. Such optimizations suggest that generating saliency maps can be done in near real-time. 
Additionally, algorithmic improvements to attribution methods have emerged; for instance, the Deep SHAP approach leverages neural network structure to compute Shapley values much faster than naïve Shapley estimates. Combining these efficient techniques enables interpretability on large-scale imaging data or time-sensitive clinical workflows\cite{Chaddad2023}. The evidence of order-of-magnitude speedups and sub-second explanation times illustrates that modern explainability methods can be practically deployed at scale, supporting clinical decisions without significant latency.

Also, we can leverage large public medical imaging datasets beyond fractures to validate our approach to broader populations and anatomies. For example, the MURA dataset of musculoskeletal radiographs spans 40,561 images across multiple body regions (elbow, wrist, shoulder), with each exam labeled normal or abnormal\cite{Abedeen2023}. Such a multi-region dataset exposes the model to varied bone anatomies and pathologies, helping assess generalization. In addition, extensive chest X-ray repositories like MIMIC-CXR, CheXpert, NIH ChestX-ray14, PadChest, and VinDr-CXR provide thousands of radiographs from diverse hospitals and patient demographics\cite{doi:10.1126/sciadv.adq0305}. These datasets introduce heterogeneity in imaging protocols (different equipment and settings) and patient profiles (age, ethnicity, clinical conditions), which would challenge our model beyond the specific fracture domain—for instance, the GRAZPEDWRI-DX collection of over 20,000 pediatric wrist X-rays offers a younger patient cohort and localized anatomy, complementing adult fracture data. By evaluating such varied datasets, we can demonstrate that our method generalizes across different populations, imaging techniques, and anatomical complexities, not just the original Bone Fracture X-ray set\cite{Abedeen2023,doi:10.1126/sciadv.adq0305}.

  In summary, Adversarial robustness and interpretability often go hand in hand: improving robustness produces interpretable maps that better align with human-understandable features. This is valuable in medical contexts because understanding the “why” behind a prediction is as important as the prediction’s accuracy.

\section{Conclusion \& Future Works}
\label{sec:conclusion}

In this work, we studied the relationship between adversarial robustness and interpretability of deep neural networks for fracture detection in X-ray images. We fine-tuned robust, pre-trained models on a diverse Bone Fracture Multi-Region X-ray dataset. The experiments demonstrated that models with better adversarial training explainable interpretability maps—saliency, Occlusion, DeepLIFT, and Integrated Gradient methods—reflect clinical reasoning. The robust models were seen to attend to relevant anatomical structures.
 
Looking ahead, several directions stem from our findings. Incorporation of additional metadata like imaging protocols, patient demographics, and clinical history can improve diagnostic accuracy and contextual relevance of interpretability maps. Using higher resolution and 3D imaging data also allows better insight into anatomical structures and more accurate localization of fracture regions. While vision transformers show promise, CNNs have long been the backbone of medical image analysis due to their robust performance, and we plan to integrate ViT-based experiments in future work to further enrich our findings. In addition, a systematic exploration of interpretability and robustness's relationship is warranted; future studies can examine how different interpretability techniques, including attention mechanisms, directly influence a model's vulnerability to adversarial attacks.

\section*{Acknowledgment}

The authors sincerely thank Dr. Ali Ghozatfar for invaluable insights and thoughtful discussions that helped develop this work.

{
    \small
    \bibliographystyle{ieeenat_fullname}
    \bibliography{main}
}


\end{document}